\documentclass[letterpaper, 10 pt, conference]{ieeeconf}  % Comment this line out if you need a4paper

\IEEEoverridecommandlockouts                              % This command is only needed if
\usepackage{algorithm, algpseudocode}
\usepackage{bm}
\usepackage{cite}
\usepackage{graphicx}
\usepackage{multirow}
\usepackage{tabularx}
\usepackage{amsmath}
\usepackage{mathrsfs}
\usepackage[super]{nth}
\usepackage{flushend}
\def\ME{ViT-HGR}
\title{\LARGE \bf \ME: Vision Transformer-based Hand Gesture Recognition from High Density Surface EMG Signals*}

\author{Mansooreh Montazerin$^{1}$, Soheil Zabihi$^{1}$, Elahe Rahimian$^{2}$, Arash Mohammadi$^{1,2}$, and Farnoosh Naderkhani$^{2}$\thanks{*This Project was partially supported by the Department of National Defence's Innovation for Defence Excellence and Security (IDEAS) program, Canada.}% <-%
\thanks{$^{1}$ Department of Electrical and Computer Engineering, Concordat University, Montreal, Canada}
\thanks{$^{1}$Concordia Institute for Information System Engineering (CITIES), Concordat University, Montreal, Canada}
}
\def\wsize {window\_size}
\def\Z{\bm{Z}}
\def\z{\bm{z}}
\def\V{\bm{V}}

\def\Q{\bm{Q}}
\def\K{\bm{K}}
\def\E{\bm{E}}

\def\x{{\mathbf x}}

\def\xp{\x^{\p}}
\def\N{N}
\def\p{p}

\def\dh{d_h}
\newcommand{\rom}[1]
{\MakeUppercase{\romannumeral #1}}

\begin{document}

\maketitle
\thispagestyle{empty}
\pagestyle{empty}
%217916829
%OOOOOOOOOOOOOOOOOOOOOOOOOOOOOOOOOOOOOOOOOOOOOOOOOOOOOOOOOOOOOO
\begin{abstract}
%OOOOOOOOOOOOOOOOOOOOOOOOOOOOOOOOOOOOOOOOOOOOOOOOOOOOOOOOOOOOOO
Recently, there has been a surge of significant interest on application of Deep Learning (DL) models to autonomously perform hand gesture recognition using surface Electromyogram (sEMG) signals. DL models are, however,  mainly designed to be applied on sparse sEMG signals. Furthermore,  due to their complex structure, typically, we are faced with memory constraints; require large training times and a large number of training samples, and; there is the need to resort to data augmentation and/or transfer learning. In this paper, for the first time (to the best of our knowledge), we investigate and design a Vision Transformer (ViT) based architecture to perform hand gesture recognition from High Density (HD-sEMG) signals. Intuitively speaking, we capitalize on the recent breakthrough role of the transformer architecture in  tackling different complex  problems together with its potential for employing more input parallelization via its attention mechanism.
%As direct application of ViT to HD-sEMG is not straightforward, a particular signal processing step is developed to convert the HD-sEMG signals to a specific format that is compatible with ViTs.
The proposed Vision Transformer-based Hand Gesture Recognition (\ME)~framework can overcome the aforementioned training time problems and can accurately classify a large number of hand gestures from scratch without any need for data augmentation and/or transfer learning. The efficiency of the proposed \ME~framework is evaluated using a recently-released HD-sEMG dataset consisting of $65$ isometric hand gestures. Our experiments with $64$-sample ($31.25~ms$) window size yield average test accuracy of $84.62 \pm3.07$\%, where only $78,210$ number of parameters is utilized. %These initial results illustrate superiority of transformer architectures for learning from HD-sEMG signals.
The compact structure of the proposed ViT-based \ME~framework (i.e., having significantly reduced number of trainable parameters) shows great potentials for its practical application for prosthetic control.
%\indent \textit{Clinical relevance}
%OOOOOOOOOOOOOOOOOOOOOOOOOOOOOOOOOOOOOOOOOOOOOOOOOOOOOOOOOOOOOO
\end{abstract}
%OOOOOOOOOOOOOOOOOOOOOOOOOOOOOOOOOOOOOOOOOOOOOOOOOOOOOOOOOOOOOO
%
%\begin{keywords}
%	Attention Mechanism, High-Density surface Electromyogram, Hand Gesture Recognition, Transformer.
%\end{keywords}

%%%%%%%%%%%%%%%%%%%%%%%%%%%%%%%%%%%%%%%%%%%%%
\section{Introduction}
%%%%%%%%%%%%%%%%%%%%%%%%%%%%%%%%%%%%%%%%%%%%%

%Why classification of sEMG dataset matters
Thanks to the recent evolution in the field of Artificial Intelligence (AI), specifically Deep Neural Networks (DNNs), significant advancements are expected on development of highly functional hand prostheses for upper limb amputees. Generally speaking, such advanced prosthesis systems are, typically, designed using surface Electromyogram (sEMG) signals~\cite{pro, pro1, Dario2, Dario:neuro, TNSRE_Elahe}, representing action potentials of the muscle fibers\cite{P2_embc}. The sEMG signals, after passing through a pre-processing stage, could be a valuable input for DNN architectures to perform different tasks including but not limited to motor control, prosthetic device control, and/or hand motion classification. Researchers are, therefore, turning their attention to development of DL-based Human Machine Interface (HMI) algorithms using sEMG signals to design more accurate and more efficient myoelectric prosthesis control systems. The ultimate goal is improving quality of life of individuals suffering from  amputated limbs.

%Why HD sEMG?
%Use more recent papers, make sure to have citations to EMBC and EMBS papers
The sEMG signals are generally classified into two main categories, i.e., sparse and high-density~\cite{HD-EMG, sparse, HD-EMG1}. Both of these multi-channel signals are acquired from a grid of electrodes placed on stump muscles to record the electrical activities and temporal information of the muscle tissue. Although using a large number of electrodes makes the computational process challenging, there has been a surge of recent interest in the use of High-Density sEMG (HD-sEMG) signals. More specifically, HD-sEMG signals are capable of recording both temporal and spatial information of the whole muscle~\cite{P1_embc}. Furthermore, by placing a large number of electrodes on the recording area, HD-sEMG provides noticeably high resolution signals compared to those obtained from sparse electrodes. This particular method improves the sparsity of the information that is collected from the muscle fibers, which results in improved classification efficiency. Moreover, these signals are independent of the exact position of the electrodes~\cite{HD adv1,HD adv2}, therefore, small variations in the position of the 2-Dimensional (2D) grid does not considerably impact the quality of the signals. Thus, development of advanced processing/learning models based on HD-sEMG signals is of significant importance to perform hand gesture recognition with high accuracy for design of myoelectric prosthesis.

\vspace{.05in}
\noindent
\textbf{\textit{Literature Review:}} Among the most prevalent approaches adopted for the task of hand gesture recognition, we can refer to the traditional Machine Learning (ML) methods such as Linear Discriminator Analysis (LDA) and Support Vector Machines (SVMs). These methods can achieve reasonable accuracy only when dealing with small data sets with a limited number of classes. However, when it comes to large data sets, most of conventional ML algorithms fail to perform at the same performance level. For such scenarios, DNNs are attractive alternatives to perform the recognition tasks with high accuracy. Convolutional Neural Networks (CNN)~\cite{CNN2, Wei2017, CNN, CNN1}, for instance, have been widely used for classification of sEMG signals by converting them into 2D images. However, such an approach is not always practical because it only considers the spatial domain of sEMG signals, ignoring their sequential nature. To jointly incorporate temporal and spatial characteristics of multi-channel sEMG signals, several hybrid models, that combine CNNs with Recurrent Neural Networks (RNN), have been proposed~\cite{PLOS2018, JMRR_Elahe}. For instance, authors in~\cite{PLOS2018} translated raw sEMG signals using six sEMG image representation techniques, which are then provided as input to a hybrid CNN-RNN model. The outcomes of~\cite{PLOS2018} eventually show that the accuracy of image classification is highly dependent on the quality of the constructed image, making the key question on ``\textit{What would be the best procedure for producing images from sEMG signals?}'' still not fully resolved~\cite{Hilbert2019}. In spite of their remarkable contributions to sequence modelling, hybrid CNN-RNN models fail to pass the entire computations for training to parallel blocks because of their innately sequential structure~\cite{Vaswani}. This raises some issues such as memory constraints and longer training times when working with large sequences. The paper aims to address this issue via incorporation of transformer architectures.

\vspace{.05in}
\noindent
\textbf{\textit{Contributions:}} In this paper, for the first time (to the best of our knowledge), we investigate and design a Transformer-based architecture~\cite{Vaswani} to perform hand gesture recognition from HD-sEMG signals. Intuitively speaking, we capitalize on the recent breakthrough role of the Transformer architecture in  tackling different complex ML problems together with its great potential for employing more input parallelization with attention mechanism. Since HD-sEMG data sets have a 3-Dimensional (3D) structure, Vision Transformers (ViT)~\cite{ViT} can be considered as an appropriate architecture to address the challenges identified above. The proposed Vision Transformer-based Hand Gesture Recognition (\ME)~framework can overcome training time problems and evaluation accuracy that we mostly face while working with other similar networks such as the conventional ML algorithms or the more advanced DNNs such as Long Short-Term Memories (LSTMs). However, we cannot directly provide HD-sEMG signals as input to the ViTs, and particular  signal processing steps are required to modify the signal into a format that is compatible with the ViT's input. Therefore, the  proposed \ME~architecture converts the main signal into smaller portions using a specific window size and then feeds each of these portions to the ViT for further analysis. In brief, contributions of this work are as follows:
\begin{itemize}
\item A Transformer-based architecture, referred to as the \ME~framework, is designed for the first time, to the best of our knowledge, to perform hand gesture recognition from HD-sEMG signals.
\item As direct application of ViT to HD-sEMG is not possible and straightforward, a particular signal processing step is developed to convert the HD-sEMG signals to a specific format that is compatible with ViTs. In other words, the proposed ViT-based \ME~framework can learn from HD-sEMG signals rather than images.
\item The proposed \ME~framework can accurately classify a large number of hand gestures from scratch without any need for data augmentation and/or transfer learning.
\end{itemize}
To develop and evaluate the proposed \ME~framework, we used a recently-released HD-sEMG dataset consisting of $65$ isometric hand gestures and $128$ distinct channels for recording the signals~\cite{data}. Our results show superior performance of the \ME~framework compared to its counterparts illustrating reduced training time and increased testing accuracy.

The remainder of the paper is organized as follows: The utilized dataset and initial preprocessing procedure are introduced in Section~\ref{section:mater}. The proposed \ME~framework together with steps used to prepare HD-sEMG signals to be fed to the ViT architecture are discussed in Section~\ref{section:vit}. Experimental results are presented in Section~\ref{section:res}. Finally, Section~\ref{sec:conc} concludes the paper.

%OOOOOOOOOOOOOOOOOOOOOOOOOOOOOOOOOOOOOOOOOOOOOOOOOOOOOOOOOOOO
\section{Materials and Methods}\label{section:mater}
%OOOOOOOOOOOOOOOOOOOOOOOOOOOOOOOOOOOOOOOOOOOOOOOOOOOOOOOOOOOO
%===========================================================
\subsection{The Dataset}\label{sub:data}
%===========================================================
The proposed \ME~framework is developed based on a database of HD-sEMG signals obtained from $20$ subjects that performed $65$ isometric hand gestures plus a gesture that was repeated twice ($66$ gestures in total)~\cite{data}. Following initial evaluations, one subjects is disregarded because of its partial information. Two grids each comprising of $64$ electrodes ($8\times8$) were placed on the flexor and extensor muscles of the participants and they were asked to repeat each movement $5$ times with five seconds rest intervals. Signals were recorded via the Quattrocento (OT Bioelettronica, Torino, Italia) biomedical amplifier system and were sampled at $2,048$ Hz frequency. The train and test sets were not mentioned in~\cite{data}. Therefore, we perform a $5$-fold cross validation to provide the bases for fair evaluations with future works. More specifically, the dataset is presented in five repetitions, where each time one repetition is used as the test set while the remaining repetitions are used as the training set. The test repetition, therefore, is changing each time and the final result is presented based on the average accuracy for each fold.

%===========================================================
\subsection{Initial Preprocessing}\label{sub:pro}
%===========================================================
The raw dataset consists of plenty of sharp fluctuations that commonly occur in the EMG signals. Not only are these fluctuations required for an accurate gesture recognition task but they also prevent the network from training the useful information and increasing its accuracy. As a result, we filter the signals of each electrode separately with a first-order low-pass butterworth filter, which yields the positive envelope of each channel. Filtered signals are then passed through a $\mu$-law normalization given by
\begin{equation}\label{mu_law}
F(x_t) = {sign}(x_t)\frac{\ln{\big(1+ \mu |x_t|\big)}}{\ln{\big(1+ \mu \big)}},
\vspace{-.1in}
\end{equation}
where  $x_t$ represents the EMG signal for each channel and $\mu$ is the parameter to which the signal is scaled. As shown in~\cite{Icassp_Elahe}, normalizing the signals increases the discriminative power of the network.

%OOOOOOOOOOOOOOOOOOOOOOOOOOOOOOOOOOOOOOOOOOOOOOOOOOOOOOOOOOO
\section{The proposed \ME~Framework}\label{section:vit}
%OOOOOOOOOOOOOOOOOOOOOOOOOOOOOOOOOOOOOOOOOOOOOOOOOOOOOOOOOOO
%%%%%%%%%%%%%%%%%%%%%%%%%%%%%%%%%%%%%%%%%%%%
\setlength{\textfloatsep}{0pt}
\begin{figure*}[t!]
\centering
\includegraphics[scale=0.42]{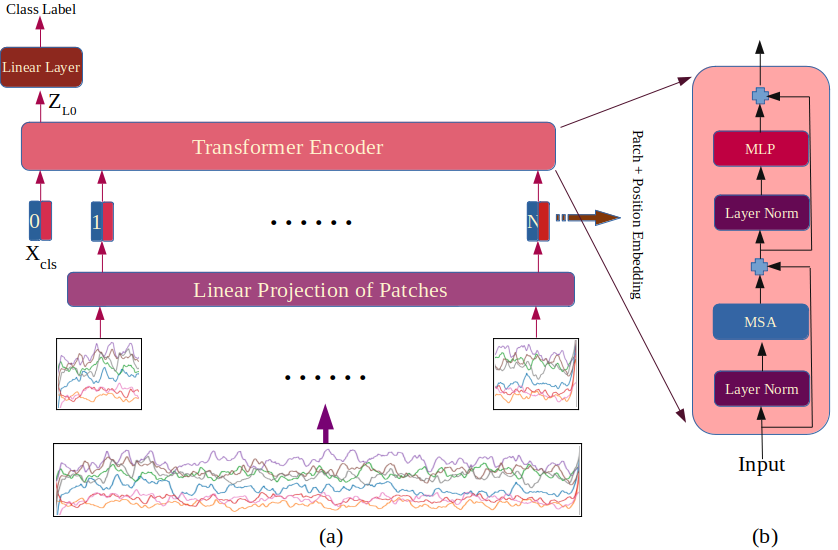}
\caption{A graphical representation of the Vision Transformer:
(a) A cropped portion of the original signal is fed to the ViT and converted to small patches. These patches go through a patch + position embedding layer and a class token is prepended to them. They are then inputted to the transformer encoder. (b) The transformer encoder that is a combination of Multi-head Self-Attention, Multilayer Perceptron and two Add\&Norm layers.\label{ViT}}
\vspace{-.2in}
\end{figure*}
%%%%%%%%%%%%%%%%%%%%%%%%%%%%%%%%%%%%%%%%%%%%
The proposed \ME~framework is implemented based on the transformers and attention mechanism~\cite{ViT}. The attention mechanism incorporated with CNNs and the LSTMs were formerly used for the hand movement classification tasks because of their proven ability to leverage the temporal information of the sEMG signals~\cite{PLOS2018}. However, in this work, we indicate that the attention framework itself is sufficient to surpass the other networks and because of our data preparation approach, there is no need for data augmentation.

The overall structure of the proposed \ME~architecture is shown in Fig.~\ref{ViT}. The ViT is designed to have a 3D image as an input, which is then reshaped to a sequence of flat 2D patches with a predetermined length. In order to convert the HD-EMG signals to the acceptable input for the ViT, we first split the whole sequence into multiple sections utilizing a sliding window of $64$ samples ($31.25$ ms considering the $2,048$ Hz sampling frequency) with a skip\_step equal to $32$ samples.
It is worth mentioning that the window length should be less than $300$ ms to satisfy the acceptable delay time~\cite{R4}, which is the real-time response required for practical myoelectric prosthetic control. Therefore, the window length for the classification purpose cannot surpass $300$ ms~\cite{R4}. Although using a larger window size potentially leads to better performance, use of shorter windows (e.g., $31.25$ ms) is preferred allowing extra necessary time for practical implementation in real scenarios.
After the windowing step, each of the (\wsize, $N_{ch}$, $N_{cv}$) sequences, where $N_{ch}$ is the number of horizontal channels and $N_{cv}$ is the number of vertical channels, are considered as the 3D input of the ViT and is divided into small square patches in the Patch Embedding block. The sequence of patches is then fed to a linear projection trainable layer ($\E$) and a class token ($\x_{cls}$) is created and put at the beginning of this sequence. The class token here is exclusively responsible for holding all the learned information during the training and is then used in the last layer to detect the gestures. In order to maintain the positional information of the signals throughout the training phase, a positional embedding vector goes through the transformer's encoder block together with each patch. The aforementioned steps provides the input ($\Z_0$) to the transformer encoder as follows:
\begin{eqnarray}
\Z_0 = [\x_{cls}; \xp_1\E; \xp_2\E;\dots; \xp_\N\E] + \E^{pos}. %\label{eq:patch}
\end{eqnarray}
The encoder block is a standard transformer encoder which is introduced by~\cite{Vaswani} and has $L$ identical layers in total. Each layer consists of two principal blocks called the ``Multi-Head Self Attention" (MSA) and the ``Feed Forward". The former is assigned for the attention mechanism while the latter acts as a position-wise Multilayer Perceptron (MLP). In the MSA block, there are $H$ number of identical heads with distinct learnable parameters operating in parallel. These heads separately accept $1/H$ of the total length of Keys ($\K$), Queries ($\Q$) and Values ($\V$) and after performing the following scaled dot-product attention
\begin{eqnarray}
\!\!\!\!\!\!SA(\Z) &\!\!\!\!=\!\!\!\!&  \text{softmax}(\frac{\Q\K^T}{\sqrt{\dh}})\V, \label{eq.4}\\
\!\!\!\!\!\!MSA(\Z) &\!\!\!\!=\!\!\!\!& [SA_1(\Z); SA_2(\Z); \dots; SA_h(\Z)]W^{MSA},\label{eq.5}
\end{eqnarray}
concatenate the outputs and transfer them to the MLP layer. The final output of the transformer after passing the MSA and MLP layers can be shown as the following matrix
\begin{eqnarray}
\Z_L = [\z_{L0}; \z_{L1}; \ldots; \z_{LN}], \label{eq.6}
\end{eqnarray}
where $N$ is the number of patches and
\begin{eqnarray}
\Z^{'}_l &=& MSA(LayerNorm(\Z_{l-1})) + \Z_{l-1},\label{eq:MSA}\\
\Z_l &=& MLP(LayerNorm(\Z^{'}_{l})) + \Z^{'}_{l}, \label{eq:MLP}
\end{eqnarray}
for $1 \leq l \leq L$. In Eq.~\eqref{eq.6}, $\z_{L0}$ is a vector holding all the useful information and is finally passed through a linear layer to produce the classification vector.

%%%%%%%%%%%%%%%%%%%%%%%%%%%%%%%%%%%%%%%%%%%%
\begin{table}[t!]
\caption{Model IDs and their parameters\label{table1}}
\vspace{-.2in}
\begin{center}
\setlength{\extrarowheight}{8pt}{
%\resizebox{\columnwidth}{!}{
\begin{tabular}{c c c}
\hline
\hline
\multicolumn{1}{c}{\textbf{Model ID}} & \textbf{MLP Size}& \textbf{Embed Dimension} \\ [0.5ex]
\hline
\rule{0pt}{3ex}
\rom{1} &  384 &192 \\

\rule{0pt}{3ex}
\rom{2}  & 96&96  \\
\rule{0pt}{3ex}
\rom{3}  &48 & 48  \\
\hline
\hline
\end{tabular}
}
\end{center}
\end{table}
%%%%%%%%%%%%%%%%%%%%%%%%%%%%%%%%%%%%%%%%%%%%
 %%%%%%%%%%%%%%%%%%%%%%%%%%%%%%%%%%%%%%%%%%%%
\begin{table*}[t!]
\caption{A Comparison of the average accuracy for each fold over $19$ participants, the overall accuracy and the overall STD for each ViT model.\label{table2}}
\vspace{-.2in}
\begin{center}
\setlength{\extrarowheight}{8pt}
\begin{tabular}{c c c c c c c c c}
\hline
\hline
\textbf{Model ID} & \textbf{Acc. F1 (\%)}& \textbf{Acc. F2} & \textbf{Acc. F3} &\textbf{Acc. F4} & \textbf{Acc. F5} &\textbf{Avg. Acc.} & \textbf{STD (\%)} & \textbf{\# Parameters}
\\
\hline
\rom{1}  &75.92& 87.79 & 88.47 & 87.71 & 83.22 & \textbf{84.62}  &  3.07   &   340,866
\\
\rom{2}  & 75.21 & 87.34 & 88 & 87.88 & 82.78 &  \textbf{84.24}  &    3.14  &   78,210  \\
\rom{3}  &   73.92 & 87.09 & 87.56 & 87.09 & 81.65 & \textbf{83.46} &  3.17   &  25,314   \\ \hline
\hline
\end{tabular}
\end{center}
\vspace{-.3in}
\end{table*}
%%%%%%%%%%%%%%%%%%%%%%%%%%%%%%%%%%%%%%%%%%%%
%%%%%%%%%%%%%%%%%%%%%%%%%%%%%%%%%%%%%%%%%%%%
\begin{table}[t!]
\caption{A Comparison of the average accuracy for each repetition over $19$ participants, the overall accuracy and the overall STD for the LDA model.\label{table3}}
\vspace{-.33in}
\begin{center}
\setlength{\extrarowheight}{8pt}
\begin{tabular}{c c c c c c c}
\hline
\hline
\textbf{Acc. F1 (\%)} \!\!\!\!&\!\!\!\! \textbf{Acc. F2} \!\!\!\!&\!\!\!\! \textbf{Acc. F3} \!\!\!\!&\!\!\!\! \textbf{Acc. F4} \!\!\!\!&\!\!\!\! \textbf{Acc. F5} \!\!\!\!&\!\!\!\! \textbf{Avg. Acc.} \!\!\!\!&\!\!\!\! \textbf{STD (\%)} \\
\hline
82.58 & 69.65 & 84.21 & 82.74 & 75.27 &\textbf{78.89} & 11.15   \\ \hline
\hline
\end{tabular}
\end{center}
\vspace{-.1in}
\end{table}
%%%%%%%%%%%%%%%%%%%%%%%%%%%%%%%%%%%%%%%%%%%%
%OOOOOOOOOOOOOOOOOOOOOOOOOOOOOOOOOOOOOOOOOOOOOOOOOOOOOOOOOOO
\section{Experiments and Results}\label{section:res}
%OOOOOOOOOOOOOOOOOOOOOOOOOOOOOOOOOOOOOOOOOOOOOOOOOOOOOOOOOOO
We evaluate our proposed framework on all the $65$ various hand gestures of the dataset and considered $3$ distinct models, in which MLP size and the Embedding dimension are different. For each model a window size of $64$ samples ($31.25$ ms) is tested to assess the impact of increasing the window size on performance of ViTs. The patch size, models' depth and the number of heads in all of the models are set to $(4,4)$, $1$ , $12$ respectively. Adam optimization method is deployed with ($\beta_{1}$ , $\beta_{2}$) equal to $(0.9 , 0.999)$, with the learning rate of $0.0001$ and the weight decay of $0.001$. We fix the batch size and the number of epochs to $128$ and $30$ respectively and use the Cross-entropy loss function for calculating the models' performance. The Model IDs and their corresponding parameters are presented in Table~\ref{table1}.

\begin{figure}[t!]
\centering
\includegraphics[scale=0.3]{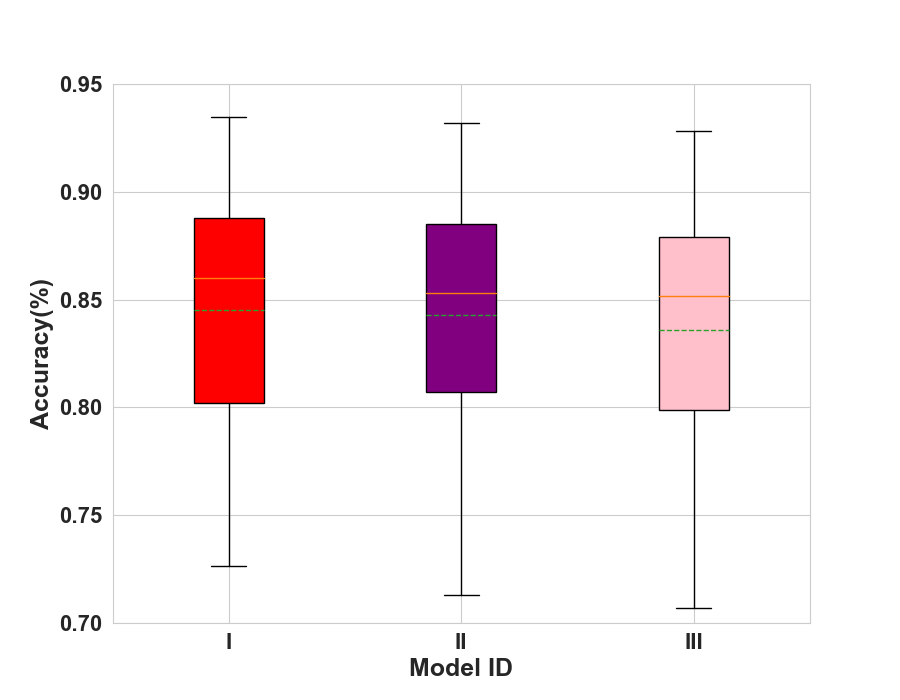}
\vspace{-.15in}
\caption{Accuracy boxplots of 3 different models of the \ME~framework. Each boxplot represents the Interquartile Range for 19 subjects. The accuracy for each subject is the average accuracy after performing 5-fold cross validation. \label{box}}
\end{figure}

Table~\ref{table2} demonstrates the average accuracy over $19$ subjects for each repetition, the average accuracy after performing 5-fold cross validation (i.e., Acc. F1 to Acc. F5) and the corresponding Standard Deviation (STD) for each model. It also shows how many parameters are trained for each specific model. As can be seen, the highest accuracy, in general, pertains to the \nth{3} repetition and the lowest to the \nth{1} one. The average accuracy rises by $0.38\%$ from model \rom{1} to model \rom{2} although the number of parameters in the former is roughly 4 times as large as that in the latter. This indicates that to obtain decent accuracy in ViTs, there is no need to increase the number of parameters and hence the complexity and the training time when this accuracy is achieved with almost $78,000$ parameters. The STD also decreases by a minimal amount when increasing the number of parameters, leading to a reasonable trade-off between the number of parameters on the one hand and the acquired accuracy and STD on the other hand. Fig.~\ref{box} visualizes the results from Table~\ref{table2}.

%%%%%%%%%%%%%%%%%%%%%%%%%%%%%%%%%%%%%%%%%%%%%
%%%%%%%%%%%%%%%%%%%%%%%%%%%%%%%%%%%%%%%%%%%%%

For comparison purposes, the proposed ViT-based framework is evaluated against the LDA approach, which is among the most popular conventional ML algorithms that has been widely used for sEMG gesture recognition. We should mention that as the utilized dataset has been released very recently, there are only couple of other works~\cite{rami, Atashzar} developed based on this dataset. Reference~\cite{rami} focused on the same task as our work but used  traditional ML methods such as LDA. The test-train split, however, is not mentioned in Reference~\cite{rami} rendering direct comparison inapplicable. Reference~\cite{Atashzar}, on the other hand,  only focused on the dynamic and transient phase of gesture movements when the signals are not stabilized or plateaued, which is a different task as this work. Consequently, to have  a fair comparison, we have implemented an LDA method similar to that of~\cite{rami} based on the same setting as our proposed ViT framework. Five key features for classification of sEMG signals including Mean Absolute Value (MAV), the number of Zero Crossings (ZC), Waveform Length (WL), Root Mean Square (RMS) and Slope Sign Change (SSC) along with four auto regressive coefficients of each cropped window are fed to the LDA algorithm~\cite{feature,rami}. In Table~\ref{table3}, the above-mentioned results obtained from the LDA model are presented. Evidently, the average accuracy in the ViT is around $5\%$ bigger and STD is around $8\%$ smaller than that in the LDA, which highlights the great power of ViTs in solving HD-sEMG hand movement classification problems. Furthermore, the signal processing + training time for both \ME and LDA and for each repetition of one subject is measured. This time is $168$ seconds for the \ME~model \rom{1} and $367$ seconds for the LDA, which means for achieving the total average accuracy over all the $19$ subjects, we require $4.4$ hours while this will be $9.6$ hours for the LDA.

%OOOOOOOOOOOOOOOOOOOOOOOOOOOOOOOOOOOOOOOOOOOOOOOOOOOOOOOOOOO
\section{Conclusions} \label{sec:conc}
%OOOOOOOOOOOOOOOOOOOOOOOOOOOOOOOOOOOOOOOOOOOOOOOOOOOOOOOOOOO
In this paper, for the first time (to the best of our knowledge), we introduced a vision transformer-based framework, referred to as the \ME, for application on HD-sEMG signals for the task of hand gesture classification.  To implement the \ME~framework, we capitalize on the recent breakthrough of Transformers in  different ML domains and their potentials for employing more input parallelization, therefore, reducing complexity of the underlying model. As direct application of ViT to HD-sEMG is not straightforward, a particular signal processing step is developed to convert the HD-sEMG signals to a specific format that is compatible with ViTs. The proposed \ME~framework can overcome the training time problems associated with recurrent networks and can accurately classify a large number of hand gestures from scratch without any need for data augmentation and/or transfer learning.
By comparing the test accuracy associated with three unique variants of the \ME~network, we showed that it could reach average accuracy of  $84.62 \%$ (for $65$ gestures over $19$ participants) with no more than $78,000$ parameters. The average accuracy for the LDA is $8\%$ lower and its signal processing + training time is more than twice that of the \ME. This illustrates potentials of the proposed \ME~framework to act as a feasible substitute for LDAs in hand gesture recognition tasks. This is because the ViTs are more straightforward to be implemented on HD-sEMG data sets with no need for any additional feature extraction calculations and it also takes far less training time, which is a significantly critical issue when working with large data sets. Our primary focus in this paper was on $64$-sample portions of the flexor signals because our purpose was to assess the proposed network's performance on small patterns of the HD-sEMG dataset. In the future, we aim to extend the number of channels to evaluate its impact on the framework's efficacy.


\begin{thebibliography}{99}

%1
\bibitem{Dario:neuro}
D. Farina, A. Mohammadi, T. Adali, N. V. Thakor, and K. N. Plataniotis,
\newblock ``Signal Processing for Neurorehabilitation and Assistive Technologies''
\newblock {\em IEEE Signal Processing Magazine,} vol. 38, no. 4, pp. 5-7, 2021.

%2
\bibitem{pro}
    S. Tam, M. Boukadoum, A. Campeau-Lecours, and B. Gosselin,
\newblock ``Intuitive Real-time Control strategy for High-density Myoelectric Hand Prosthesis Using Deep and Transfer Learning.''
\newblock {\em  Scientific Reports}, vol. 11, no. 1, pp. 1-14, 2021.

%3
\bibitem{pro1}
    J. Vogel, and A. Hagengruber,
\newblock ``An sEMG-based Interface to Give People with Severe Muscular Atrophy Control over Assistive Devices.''
\newblock {\em IEEE Int. Conf. of the Engineering in Medicine and Biology Society (EMBC)}, pp. 2136-2141, 2018.

%4
\bibitem{Dario2}
M. Zia ur Rehman, \text{et al.},
\newblock ``Stacked Sparse Autoencoders for EMG-based Classification of Hand Motions: A Comparative Multi Day Analyses between Surface and Intramuscular EMG,''
\newblock {\em Appl. Sci.}, vol. 8, p. 1126,  2018.

\bibitem{TNSRE_Elahe}
E. Rahimian, S. Zabihi, A. Asif, D. Farina, S.F. Atashzar, A. Mohammadi,
\newblock ``FS-HGR: Few-shot Learning for Hand Gesture Recognition via ElectroMyography,''
\newblock {\em IEEE Trans. Neural Syst. Rehabil. Eng.}, 2021.

\bibitem{P2_embc}
P. Koch, M. Dreier, M. Maass, M. Böhme, H. Phan, A. and Mertins,
\newblock ``A Recurrent Neural Network for Hand Gesture Recognition Based on Accelerometer Data,''
\newblock {\em  IEEE Int. Conf. of the  Engineering in Medicine and Biology Society (EMBC)}, pp. 5088-5091, 2019.

%5
\bibitem{HD-EMG}
D. Bai, S. Chen, and J. Yang,
\newblock ``Upper Arm Motion High-density sEMG Recognition Optimization Based on Spatial and Time-frequency Domain Features,''
\newblock {\em Journal of Healthcare Engineering}, 2019.

%6
\bibitem{sparse}
I. Ketykó, F. Kovács, and K. Z. Varga,
\newblock ``Domain Adaptation for sEMG-based Gesture Recognition with Recurrent Neural Networks,''
\newblock {\em IEEE Int. Joint Conf. Neural Networks (IJCNN)}, pp. 1-7, 2019.

%7
\bibitem{HD-EMG1}
U. Kuruganti, A. Pradhan, and J. Toner,
\newblock ``High-Density Electromyography Provides Improved Understanding of Muscle Function for Those with Amputation,''
\newblock {\em Frontiers in Medical Technology}, vol. 41, 2021.

\bibitem{P1_embc}
M. Houston, A. Wu, Y. and Zhang,
\newblock ``Optimizing Input for Gesture Recognition using Convolutional Networks on HD-sEMG Instantaneous Images,''
\newblock {\em IEEE Int. Conf. of the Engineering in Medicine and Biology Society (EMBC)}, pp. 6539-6542, 2021.

%8
\bibitem{HD adv1}
J. Chen, S. Bi, G. Zhang, and G. Cao,
\newblock ``High-density Surface EMG-based Gesture Recognition Using a 3D Convolutional Neural Network.,''
\newblock {\em Sensors}, vol. 20, no. 4, p. 1201, 2020.

%9
\bibitem{HD adv2}
X. Jiang, X. Liu, J. Fan, et al.,
\newblock ``Open Access Dataset, Toolbox and Benchmark Processing Results of High-Density Surface Electromyogram Recordings.,''
\newblock {\em IEEE Trans. Neural Syst. Rehabil. Eng.}, vol. 29, pp. 1035-1046, 2021.

%10
\bibitem{CNN}
S. Shen, K. Gu, X. R. Chen, M. Yang, and R. C. Wang,
\newblock ``Movements Classification of Multi-channel sEMG Based on CNN and Stacking Ensemble Learning,''
\newblock {\em IEEE Access}, vol. 7, pp. 137489-137500, 2019.

%11
\bibitem{CNN1}
A. Gautam, M. Panwar, A. Wankhede, S. P. Arjunan,  G. R. Naik, A. Acharyya, and D. K. Kumar,
\newblock ``Locomo-net: A Low-complex Deep Learning Framework for sEMG-based Hand Movement Recognition for Prosthetic control,''
\newblock {\em IEEE J. Transl. Eng. Health Med.}, vol. 8, pp. 1-12, 2020.

%12
\bibitem{CNN2}
T. Triwiyanto, I. P. A. Pawana, and M. H Purnomo,
\newblock ``An Improved Performance of Deep Learning Based on Convolution Neural Network to Classify The Hand Motion by Evaluating Hyper Parameter,''
\newblock {\em IEEE Trans. Neural Syst. Rehabil. Eng.}, vol. 28, no. 7, pp. 1678-1688, 2020

%13
\bibitem{Wei2017}
W. Wei, \textit{et al.},
\newblock ``A Multi-stream Convolutional Neural Network for sEMG-based Gesture Recognition in Muscle-computer Interface,''
\newblock {\em Pattern Recognition Letters}, 2017.

%14
\bibitem{PLOS2018}
Y. Hu, Y. Wong, W. Wei, Y. Du, M. Kankanhalli, and W. Geng,
\newblock ``A Novel Attention-based Hybrid CNN-RNN Architecture for sEMG-based Gesture Recognition,''
\newblock {\em PloS One}, vol. 13, no. 10, 2018.

%15
\bibitem{JMRR_Elahe}
E. Rahimian, S. Zabihi, S. F. Atashzar, A. Asif, A. Mohammadi,
\newblock ``Surface EMG-Based Hand Gesture Recognition via Hybrid and Dilated Deep Neural Network Architectures for Neurorobotic Prostheses,''
\newblock {\em Journal of Medical Robotics Research}, pp. 1-12, 2020.

%16
\bibitem{Hilbert2019}
P. Tsinganos, B. Cornelis, J. Cornelis, B. Jansen, and A. Skodras,
\newblock ``A Hilbert Curve Based Representation of sEMG Signals for Gesture Recognition,''
\newblock {\em International Conference on Systems, Signals and Image Processing (IWSSIP)}, pp. 201-206, 2019.

%17
\bibitem{Vaswani}
A. Vaswani, N. Shazeer, J. Uszkoreit, L. Jones, A. Gomez N., L. Kaiser, and I. Polosukhin,
\newblock ``Attention is All You Need,''
\newblock {\em Advances in Neural Information Processing Systems}, pp. 5998-60, 2017.

%18
\bibitem{ViT}
A. Dosovitskiy, L. Beyer, A. Kolesnikov, et al.,
\newblock ``An Image is Worth 16x16 Words: Transformers for Image Recognition at Scale,''
\newblock {\em  arXiv preprint arXiv:2010.11929}, 2020.

%19
\bibitem{data}
N. Malešević, A. Olsson, P. Sager, et al.,
\newblock ``A Database of High-density surface Electromyogram Signals Comprising 65 Isometric Hand Gestures,''
\newblock {\em Scientific Data,}  vol. 8, no. 1, pp. 1-10, 2021.

%20
\bibitem{Icassp_Elahe}
E. Rahimian, S. Zabihi, F. Atashzar, A. Asif, A. Mohammadi,
\newblock ``XceptionTime: Independent Time-Window XceptionTime Architecture for Hand Gesture Classification,''
\newblock {\em IEEE Int. Conf. on Acoustics, Speech, and Signal Processing (ICASSP)}, 2020.

\bibitem{R4}
B. Hudgins, P. Parker, and R.N. Scott,
\newblock ``A New Strategy for Multifunction Myoelectric Control,''
\newblock {\em IEEE Trans. Biomed. Eng.}  vol. 40, no. 1, pp. 82-94, 1993.


%21
\bibitem{rami}
R. N. Khushaba, and K. Nazarpour,
\newblock ``Decoding HD-EMG Signals for Myoelectric Control - How Small Can the Analysis Window Size be?,''
\newblock {\em IEEE Robotics \& Automation Letters,} vol. 6, no. 4, 2021.

%22
\bibitem{feature}
    M. Atzori, and H. Müller,
\newblock ``PaWFE: Fast Signal Feature Extraction Using Parallel Time Windows.''
\newblock {\em  Frontiers in Neurorobotics,} vol. 13, p. 74, 2019.

% %23
 \bibitem{Atashzar}
 T. Sun, Q. Hu, P. Gulati, and S.F. Atashzar,
 \newblock ``Temporal Dilation of Deep LSTM for Agile Decoding of sEMG: Application in Prediction of Upper-limb Motor Intention in NeuroRobotics.,''
 \newblock {\em IEEE Robotics and Automation Letters},  2021.

\end{thebibliography}
\end{document}